\documentclass[letterpaper, 10 pt, conference]{ieeeconf}  

\IEEEoverridecommandlockouts                              

\usepackage{amsmath} 
\usepackage{flushend}

\usepackage{amssymb}  
\usepackage{graphicx}
\usepackage{subcaption}
\usepackage{float}
\usepackage{placeins}
\usepackage{xcolor}
\usepackage[most]{tcolorbox}
\usepackage[table,dvipsnames]{xcolor}
\usepackage{colortbl}

\usepackage[colorlinks=true, citecolor=magenta]{hyperref}

\newcommand{\framework}{\textsc{AlphaAdj}}

\usepackage{color}
\usepackage[dvipsnames]{xcolor}
\usepackage{colortbl}
\usepackage{xcolor}
\usepackage{mathrsfs}
\usepackage{tikz}
\usetikzlibrary{positioning}
\usepackage{xspace}
\usepackage{multicol,multirow}

\usepackage{amsthm}
\usepackage[outline]{contour}
\usepackage{soul}
\usepackage[]{mdframed}
\usepackage{threeparttable}
\usepackage{pifont}
\usepackage{wrapfig}
\usepackage{amsmath, amsfonts, amssymb}   
\usepackage{mathtools}
\usepackage{pifont}%
\let\labelindent\relax
\usepackage{enumitem}
\usepackage{booktabs}
\usepackage{float}
\usepackage[linesnumbered,ruled,vlined]{algorithm2e}
\usepackage{bm}
\usepackage{subcaption}
\usepackage{makecell}
\usepackage[font=small]{caption}

\usepackage{txfonts}

\title{\LARGE \bf

Semantically Adaptive Control Barrier Functions with Vision-Language Models for Real-Time Robot Navigation

}

\author{Jeffrey Chen, Rohan Chandra\\
{\small \href{https://saferobotnav.github.io/AlphaAdj/}{ Supplementary Material, Videos at the following anonymous website: \textbf{saferobotnav.github.io/AlphaAdj}}} \\
\thanks{Authors are with the Department of Computer Science, University of Virginia. {\tt\small \{fyy2ws, aar8xx\}@virginia.edu}}
}

\begin{document}

\maketitle
\thispagestyle{empty}
\pagestyle{empty}

\bstctlcite{IEEEexample:BSTcontrol}


\begin{abstract}
Robots operating in dynamic, unstructured environments must maintain safety without becoming unnecessarily conservative. While control barrier functions (CBFs) provide a principled mechanism for collision avoidance, their behavior is typically determined from geometric state alone and therefore cannot distinguish between semantically different situations with similar physical configurations. We present \framework{}, a vision-to-control navigation framework that uses egocentric RGB observations and a vision-language model (VLM) to adapt CBF conservativeness according to visual semantic context. The VLM produces a scalar contextual caution signal that dynamically modulates the CBF parameter $\alpha$, while a geometry- and speed-dependent cap bounds admissible aggressiveness. To support real-time operation despite VLM latency, inference runs asynchronously with the control loop, and the semantic signal is extracted directly from token-level output probabilities rather than full autoregressive text generation. We evaluate \framework{} across matched-geometry, dynamic-human, and cluttered navigation scenarios, as well as on a physical Clearpath Jackal. When obstacle geometry is held fixed, \framework{} allocates 0.405 m more clearance to a human than to a benign object, while geometry-only and non-visual adaptive CBF baselines produce no semantic differentiation. This additional behavioral reserve improves robustness to unexpected human motion, including cluttered conditions in which all evaluated geometric baselines collide while \framework{} remains collision-free. The system maintains a 0.0167 s control period with 0.158 s mean VLM round-trip latency, demonstrating that semantic visual reasoning can complement geometric safety control without entering the critical control path.

\end{abstract}



\section{Introduction}

Autonomous robots increasingly operate in dynamic, crowded, and unstructured environments. In such settings, robots must maintain safety without becoming unnecessarily conservative. Control barrier functions (CBFs) provide a principled mechanism for enforcing safety constraints during navigation~\cite{ames2016_cbf, ames2019_cbf}. In practice, however, the behavior induced by a CBF depends strongly on parameters that determine how conservatively the robot responds to nearby obstacles~\cite{garg2023advancestheorycontrolbarrier}. A single fixed setting can therefore be poorly suited to environments whose demands may vary substantially over time. Conservative settings provide additional margin but may produce unnecessary detours, while permissive settings enable more efficient to-goal motion but leave less behavioral reserve when more caution is warranted. 

Many existing adaptive safety methods~\cite{margolis2021,gouru2024,zinage2024} reason primarily on the geometric and dynamic state of the surrounding environment, as such information is essential for collision avoidance. However, in practice, such information is not enough to fully determine how a robot should behave around an obstacle. For example, a stationary box and a person may occupy nearly identical positions, but warrant very different navigation behavior. Passing closely and efficiently by the box may be desired, while maintaining additional clearance around the person provides room for unexpected human movement and avoids undesirably close interaction. 


Recent advances in vision-language models (VLMs) and multimodal foundation models provide a promising source of this missing contextual information, as these models can infer scene-level understanding directly from visual and text input~\cite{bordes2024introductionvisionlanguagemodeling}. Recent frontier models, such as GPT-6 Astra and Claude Fable 5.1, further demonstrate increasingly capable multimodal and visual zero-shot reasoning across a broad range of previously unseen tasks~\cite{openai2026astra,anthropic2026fable51}. These capabilities suggest that semantic visual reasoning can complement geometric safety control. Rather than replacing the low-level safety controller, a VLM can provide complementary contextual information about what the robot is approaching and how cautiously the controller should respond. 

This distinction creates a fundamental limitation for controllers whose conservativeness depends only on geometry. If a robot encounters, at the exact same location, either a benign static object or a person, a geometry-only controller receives essentially the exact same information in both cases and therefore has no reason to allocate different amounts of clearance. Separately, a globally conservative controller could maintain large amounts of clearance in both scenes, but would pay an efficiency cost for the benign object case. What we seek instead is adaptive, selective conservativeness, which results in efficient navigation when additional clearance is unnecessary, while allocating greater clearance when visual context indicates that more caution is appropriate.  


Using a VLM in real-time control introduces a separate practical challenge. Modern foundation models run substantially slower than conventional robot controllers, so VLM-based visual navigation systems can suffer from high latency for real-time robot navigation~\cite{hirose2026asyncvlaasynchronousvlafast, song2024vlmsocialnavsociallyawarerobot, huang2026ticvla}. Waiting synchronously for seconds-long VLM inference calls delays robot motion and is infeasible for real mobile robot deployment. A useful architecture should instead provide a lightweight interface through which comparatively slow semantic reasoning can modulate a fast, geometric controller.

In this work, we present \framework{}, an adaptive navigation framework that uses high-level visual semantics to adapt low-level CBF conservativeness in real time. An egocentric RGB observation is processed by a VLM to produce a scalar contextual caution signal describing how strongly the robot should yield to what it observes in the current visual scene. This signal is used to dynamically adjust $\alpha$, a CBF control parameter which governs conservativeness. Geometry therefore continues to define the control envelope, while visual semantics determine how much additional behavioral reserve is used within that envelope. 

To support continuous robot movement, the VLM inference runs asynchronously with the control loop. Rather than wait for full autoregressive text generation and subsequently parsing the scalar from text, \framework{} extracts the caution signal directly from token-level output probabilities, reducing inference overhead and allowing frequent semantic updates. The low-level controller can therefore continue operating while new semantic information is computed. This design also provides a modular interface between perception and control. As multimodal foundation models continue to improve, such stronger semantic backbones can be easily incorporated without changing the underlying controller. 

\subsection{Main Contributions}
\begin{itemize}
    \item \textbf{Semantically adaptive navigation:}
    We introduce a vision-to-control framework that uses VLM-derived semantics on egocentric RGB inputs to dynamically adjust CBF conservativeness. The formulation preserves geometry as the physical safety layer, while allowing visual semantics to selectively allocate additional behavioral reserve.  

    \item \textbf{Real-time architecture:}
    We develop an asynchronous VLM-to-control architecture that decouples comparatively slow VLM inference from the low-level controller. A scalar contextual caution signal is obtained directly from next-token probabilities, while a geometry- and motion-dependent cap bounds its influence on the geometric controller. 
    
    \item \textbf{Evaluation of semantic-driven behavioral reserve:}
    We evaluate whether changing only the semantic identity of an obstacle produces different navigation behavior when the physical configurations are otherwise unchanged, and compare against geometry-only adaptation, globally conservative CBF settings, and non-visual adaptive CBF baselines. We further evaluate robustness to unexpected human motion, performance across increasing scene complexity, and transfer to a physical Clearpath Jackal. 
\end{itemize}

\section{Related Work}
\label{Related Work}

\subsection{Adaptive CBF Parameter Tuning}
\label{sec:related_adaptive_cbf}

Control Barrier Functions (CBFs) are a powerful mathematical framework used to encode robot system safety in nonlinear systems~\cite{ames2016_cbf, ames2019_cbf}. Adaptive CBF methods study how to adjust CBF parameters in order to further optimize robot performance while staying within these safe sets. Early formulations introduce adaptive CBFs that update safety certificates in response to uncertainty and changing operating conditions, improving practicality compared to static, hand-tuned choices~\cite{xiao2020adaptivecontrolbarrierfunctions}. However, rather than focus on overall system safety and performance, these earlier adaptive CBF approaches focus more on addressing the uncertainty of the dynamic model~\cite{Black_2021, Taylor_2019, Taylor_2020}. More recent work, such as OA-CBF, uses learning to choose or refine CBF parameters during operation~\cite{kim2026learning}. PACBF learns how to adapt CBF settings and checks that the chosen updates still preserve safety, demonstrating the idea on agile flight platforms such as a quadplane~\cite{kim2025adaptcontrolbarrierfunctions}. 
Complementary work considers safety constraints that themselves change over time 
and adapts CBFs in a robust way to remain safe even with modeling errors and disturbances~\cite{kim2025robustadaptivetimevaryingcontrol}. 

Our method shares the same overall goal of adapting CBF behavior online, but differs in what information drives the adaptation. Instead of adapting CBF parameters primarily from learned dynamics models or by searching over candidate parameter settings, \framework{} uses semantic information derived directly from egocentric RGB observations, using a VLM to adapt the CBF parameter $\alpha$. This allows the controller to differentiate between different situations that may be geometrically similar.


\subsection{Online Planner Parameter Tuning}
\label{sec:related_appl}

Our approach is philosophically related to Adaptive Planner Parameter Learning (APPL)~\cite{xiao2022appl}, which argues that classical navigation stacks often require extensive manual retuning to generalize across environments~\cite{xiao2022motionplanningcontrolmobile}, and that even within a single environment a fixed parameter set may perform inconsistently across different contexts~\cite{xiao2020appld}. APPL addresses this by learning a policy that adjusts planner parameters online during deployment, through signals such as demonstrations, interventions, evaluative feedback, or reinforcement learning in simulation~\cite{xiao2020appld,wang2021appli,wang2021apple,xu2021applr}.

APPL reparameterizes a classical planner using learned policies from demonstrations and feedback, whereas we adapt the conservativeness of a CBF constraint directly from visual semantic context, without requiring task specific training.

\subsection{Vision-Based Navigation}
There is increasing interest in using vision directly for control and safety. Frameworks like LivePoint~\cite{chen2025livepointfullydecentralizedsafe} successfully integrate CBFs for decentralized multi-robot navigation using only LiDAR-derived point cloud inputs, while other RGB-geometry based approaches~\cite{chen2025safersplatcontrolbarrierfunction} and~\cite{tscholl2025perceptionintegratedsafetycriticalcontrol} couple 3D scene geometry (via 3D Gaussian splatting) with CBF-based filtering to activate safety constraints more proactively. Visual-servoing techniques likewise use camera feedback directly to regulate robot motion~\cite{ahmadi2020, li2020}. These approaches demonstrate that vision can inform low-level control, but they typically rely on geometric information and provide fixed safety behavior once deployed.


More recently, VLMs have been used to incorporate higher-level semantic and contextual understanding into robot navigation. VLM-Social-Nav uses a VLM-based scoring module to compute a socially aware cost term that guides an underlying motion planner~\cite{song2024vlmsocialnavsociallyawarerobot}. CoNVOI uses VLMs to identify environmental context and select visually grounded regions that are navigable, which are then converted into reference paths for the underlying planner~\cite{sathyamoorthy2024convoicontextawarenavigationusing}. \framework{} is related in its use of semantic visual information, but differs in where that information enters the stack. Rather than generating reference trajectories, modifying costs, or producing actions, we use the VLM signal to regulate the conservativeness of a low-level CBF constraint.

There are further challenges that come with usage of large foundation models in mobile robots. Because such models can easily exceed the onboard compute budget of mobile robots, inference may require being offloaded to remote or cloud-based systems~\cite{niwa2022spatiotemporalgraphlocalizationnetworks}. Such reliance on external setups may degrade performance as communication latency between the robot and the remote system lead to delays~\cite{hirose2026asyncvlaasynchronousvlafast}. Such challenges directly motivate latency-aware adaptive vision-based control, where visual understanding is used not only to set an initial safety policy, but to continuously adjust how the controller trades off safety and progress in real time.

\section{Preliminaries}
\label{Preliminaries}

In this section, we formally introduce the navigation problem we study and the safety framework we use through relevant background on CBFs. 

We study safe navigation for a single ground robot in a cluttered environment using egocentric RGB observations. We formulate a single agent navigation scenario using the following tuple: $\langle \mathcal{X}, \mathcal{U}, \mathcal{T}, \mathcal{J}, \Omega, O \rangle,$ with discrete time $t\in\{0,\dots,T\}$. The robot state $x(t)\in\mathcal{X}\subset\mathbb{R}^n$ includes the robot position $p(t)\in\mathbb{R}^2$ and heading (yaw) $\psi(t)\in\mathbb{R}$.
The control $u(t)\in\mathcal{U}\subset\mathbb{R}^2$ is a desired planar velocity vector in the world frame. At each step, $u(t)$ is converted into differential-drive commands $(v(t),\omega(t))$ using the current heading $\psi(t)$, where $v(t)$ is the commanded forward speed and $\omega(t)$ is the commanded yaw rate. The robot follows discrete and deterministic control-affine dynamics 

\begin{equation}
x(t+1) = f(x(t)) + g(x(t))u(t)
\label{eq:affine_system}
\end{equation}

\noindent where functions $f, g$ are locally Lipschitz continuous functions. The robot receives observations $o(t)$ in the form of egocentric visual images. At each control step, the robot uses the current semantic information together with the geometric state to determine a CBF conservativeness parameter $\alpha(t)$, and then solves for a safe control input $u_\textrm{safe}$ conditioned on that $\alpha(t)$.


\subsection{Safe Control}

We formalize safety by discussing Control Barrier Functions (CBFs) \cite{ames2016_cbf}. A CBF is a scalar function, $h({x})$ that is defined over the state space of a system. We define the safe set  $\mathcal{C}$ as the set of all states for which 
$h(x) > 0$. Ensuring safety involves keeping the system state within $\mathcal{C}$ at all times; safety is guaranteed as long as the CBF is always non-negative. This is achieved by designing a control input, $u$, such that the following condition is satisfied (For notational simplicity, here and moving forward, we omit explicit time $t$ arguments when unambiguous):

\begin{equation}
\frac{d}{dt}h(x) \ge -\alpha(h(x))
\label{eq:cbfcondition}
\end{equation}
where $\frac{d}{dt}h(x)$ denotes the time derivative of $h(x)$ along system trajectories and $\alpha$ is an extended class $\mathcal{K}$ function, typically a linear or higher-order function that ensures the safety constraint is enforced with appropriate robustness.  This condition limits how quickly the robot is allowed to move toward unsafe states. When the robot is far from the safety boundary ($h(x)$ is large, positive), the condition is easily satisfied and the controller has more freedom. But as the robot approaches the boundary $h(x) = 0$, the constraint tightens, requiring $\frac{d}{dt}h(x)$ 
to be either non-negative or not too negative, which means the robot cannot get closer to the boundary quickly.  

We seek safe yet efficient navigation by dynamically adjusting CBF behavior using risk information inferred from the VLM. In particular, we adapt the CBF $\alpha$ parameter, which controls the aggressiveness of the safety constraint enforcement: smaller $\alpha$ yields more conservative behavior near obstacles, while larger $\alpha$ permits more aggressive motion. We treat the CBF conservativeness $\alpha(t)$ as a time-varying scalar $\alpha(t)\in[\alpha_{\min},\alpha_{\max}]$ calculated online based on the current RGB observation. We enforce safety by filtering nominal controls through a CBF constraint for control-affine dynamics (Eq.~\ref{eq:affine_system}). Substituting Eq.~\ref{eq:affine_system} into the CBF condition in Eq.~\ref{eq:cbfcondition} yields

\begin{equation}
\frac{\partial h}{\partial x}\bigl(f(x)+g(x)u\bigr) \ge -\alpha\!\left(h(x)\right)
\label{eq:cbfconditionfull}
\end{equation}

This inequality lower-bounds the rate of change of the barrier function $h(x)$ along system trajectories. When $\alpha$ is small, the right-hand side is less negative, which forces  $\frac{d}{dt}h(x)$ to be larger (i.e., $h$ must decrease more slowly or increase), leading to more conservative avoidance. When $\alpha$ is large, the bound becomes more negative, allowing $\frac{d}{dt}h(x)$ to take more negative values (i.e., permitting $h$ to decrease faster), which corresponds to more aggressive motion that can approach obstacles more quickly before corrective action is required. In our implementation, we use a per-obstacle CBF defined using a squared-distance safety margin.
Let $p\in\mathbb{R}^2$ denote the robot position in the plane and let $c_j\in\mathbb{R}^2$ be the center of circular obstacle $j$ with radius $R_j$. We model the robot with an effective radius $R_R$, a circular safety approximation of the robot body, and include an optional safety padding $\delta$.
Defining $R_{\text{sum},j}=R_R+R_j+\delta$, the CBF for obstacle $j$ is thus:
\begin{equation}
h_j(x) = \|p-c_j\|^2 - R_{\text{sum},j}^2.
\label{eq:hj}
\end{equation}

\noindent We first compute a nominal velocity $u_{\textrm{nom}}$ that points the robot toward the goal, with its speed capped at the robot's maximum speed $v_\textrm{max}$. We then compute a safe velocity $u_{\text{safe}}$ by solving a small quadratic program that minimizes deviation from $u_{\textrm{nom}}$ while enforcing the CBF constraint in Eq.~\ref{eq:cbfconditionfull}. The QP is formulated as:
\begin{equation}
    u_\textrm{safe} = \arg\min_{u \in \mathcal{U}} \frac{1}{2} \|u - u_{\textrm{nom}}\|^2 
    \label{eq:QP}
\end{equation}
\[
\text{subject to CBF constraints holding and} \ \|u\| \leq v_\textrm{max}.
\]

This formulation preserves the goal-seeking behavior of $u_{nom}$ as much as possible while guaranteeing collision avoidance through the CBF constraint.

\section{\framework{}: Approach}
\label{sec:approach}

We present \framework{}, a real-time framework that uses egocentric RGB to adapt the conservativeness of a CBF safety filter according to visual semantic context. At a high level, the robot continuously runs a nominal goal-seeking controller subject to geometric CBF constraints, while a vision-language model (VLM) periodically analyzes the current RGB frame and estimates how much additional caution or yielding is warranted by the scene. This semantic signal is used to adapt the CBF parameter $\alpha$, allowing the controller to allocate different levels of conservativeness when obstacle geometry is similar. 

\begin{figure}[t]
    \centering
    \includegraphics[width=0.47\textwidth]{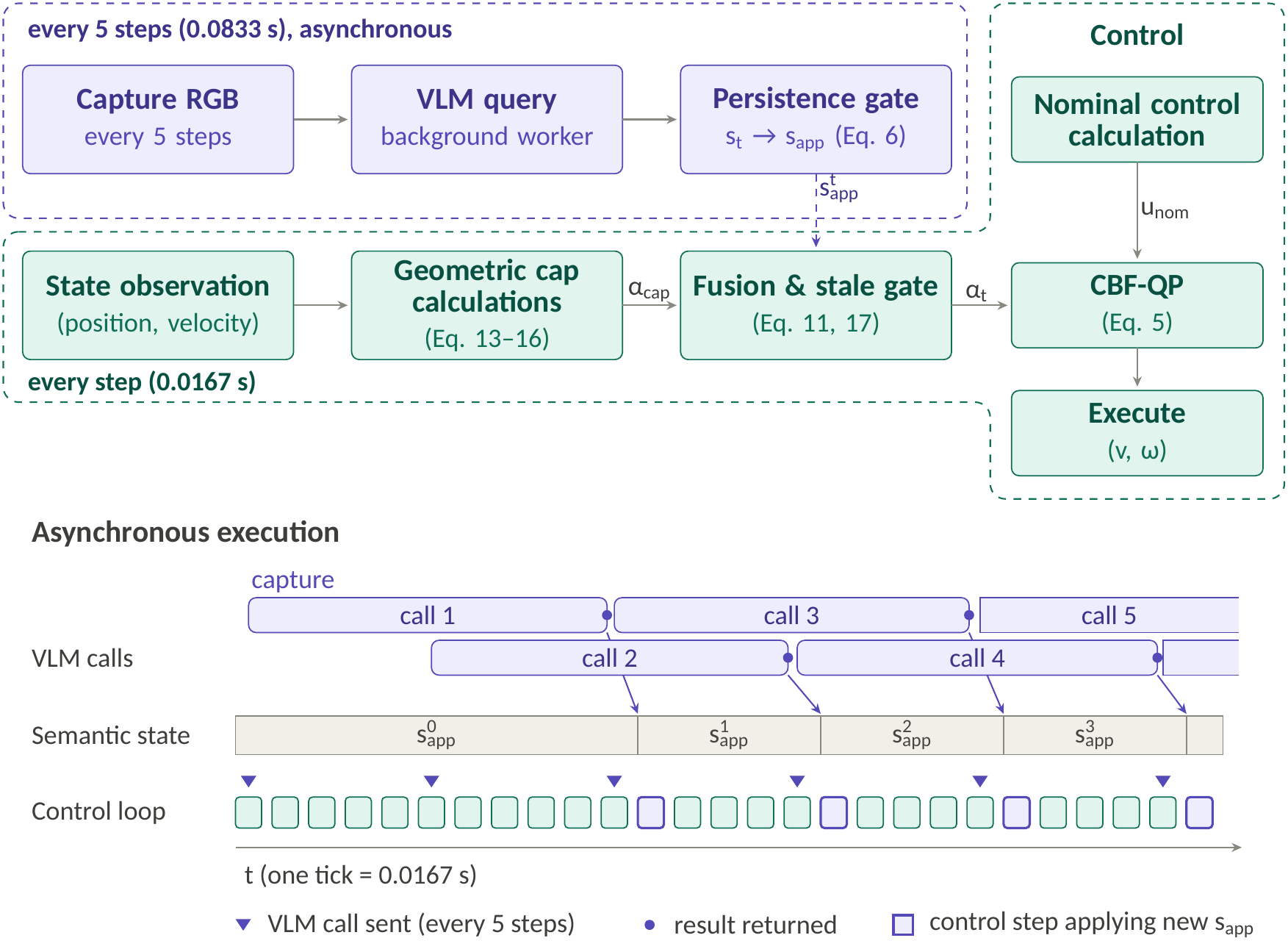}
    \caption{
    Overview of \framework{}. Egocentric RGB frames are processed asynchronously by a VLM to produce a contextual caution score $s_t$ from token-level output probabilities. The semantic signal modulates the CBF parameter $\alpha_t$, while a geometric, speed-aware cap $\alpha_{\text{cap}}$ bounds admissible aggressiveness. The low-level controller continues operating with the most recent valid semantic estimate while new VLM inference is outstanding.
    }
    \label{fig:overview}
    \vspace{-21pt}
\end{figure}

Because VLM inference is substantially slower than the low-level control loop, VLM inference is executed asynchronously while control continues at every timestep. To preserve semantic context throughout an interaction, \framework{} associates the contextual caution estimate with the geometric interaction currently most relevant to the CBF and retains that estimate while the same interaction remains active, even if the robot's viewpoint temporarily changes. We additionally constrain the semantic adaptation using a geometry and speed dependent cap, preventing faulty or delayed semantic estimates from making the controller more aggressive than permitted by the current physical state. This design yields a simple division of responsibility. Geometry bounds admissible aggressiveness, while visual semantics determine how much additional behavioral reserve should be allocated within that bound. The overall system overview is shown in Fig.~\ref{fig:overview}.

We consider a mobile robot with egocentric RGB camera sensing operating in an environment with static or dynamic obstacles. The low-level controller operates at a higher rate than VLM inference, executing every $0.0167$ seconds (60 Hz). Every 5 control steps ($0.0833$ seconds, 12 Hz), \framework{} captures an RGB frame and asynchronously queries a locally hosted VLM while the controller continues to execute. We utilize Qwen2.5-VL-7B-Instruct, giving it only the egocentric RGB image and a general scene-level semantic prompt. Geometric quantities such as obstacle distance, barrier value, and robot speed are not supplied to the VLM. 
This design intentionally separates semantic reasoning from the geometric state used by the CBF controller. The current prompt asks: 

\begin{tcolorbox}[colback=blue!5, breakable, colframe=blue!40!black, title=Server-Side VLM Instruction]
\textit{A mobile robot shares this space with whatever is visible ahead. How strongly should the robot yield and avoid imposing on, disturbing or crowding what it sees? 0 = no need to yield, 1 = yield strongly. Return strict JSON only: \{"score": $<$one decimal between 0.00 and 1.00$>$\}}
\vspace{-5pt}
\end{tcolorbox}

Additional prompt formulations and their corresponding evaluations are provided in supplementary material. 


The VLM produces a scalar contextual caution score $s_t\in[0,1]$ where larger values indicate that the current visual context warrants greater caution beyond that implied by geometric collision avoidance alone. We do not interpret $s_t$ as a calibrated probability of collision. Instead, $s_t$ provides a semantic control signal derived from information available in the RGB observation, such as whether nearby entities may move unpredictably, or whether passing them at close range is appropriate. This intentional separation allows geometric state to remain available to the low-level controller, while high level semantic information contributes contextual information that geometry alone does not encode.

\subsubsection{Token-Probability Readout}
A limitation of using VLM-generated text directly for real-time control is that the system must wait for full decoding to finish before parsing the response. Instead, \framework{} extracts the semantic score $s_t$ directly from the model's output-token probabilities. 


We constrain the beginning of the numerical model response and perform a single forward pass. Let $T(d)$ denote the set of all equivalent tokenizer IDs corresponding to a digit $d\in{0,\ldots,9}$. Given RGB image $I_t$ and prompt $P$, we define the probability mass assigned to digit $d$ as:

\begin{equation}
M_t(d)
=
\sum_{\tau\in T(d)}
p(\tau\mid I_t,P).
\label{eq:digit_mass}
\end{equation}

\framework{} then computes contextual caution score as the probability weighted expected digit value:

\begin{equation}
s_t
=
\frac{1}{10}
\frac{
\sum_{d=0}^{9} d\,M_t(d)
}{
\sum_{d=0}^{9} M_t(d)
}.
\label{eq:semantic_score}
\end{equation}

This probability based formulation retains and smooths information across multiple plausible numerical outputs, while avoiding the time delay of a complete textual response.


\subsection{Interaction-Persistent Semantic Feedback}
\label{sec:interaction_persistent}

The contextual caution score is scene-level, while the robot's egocentric viewpoint may change continuously during navigation. In particular, an avoidance maneuver may turn the camera away from an interaction that is still geometrically active and relevant. Thus treating each image independently would therefore allow useful semantic context to disappear simply because the robot has begun responding to it. We instead make the semantic signal persistent over the active geometric interaction. At each control step, we identify the obstacle currently most relevant to the CBF using the same barrier quantities that are already available to the geometric controller: 

\begin{equation}
j_t^\star=\arg\min_j h_j(x_t).
\label{eq:active_obstacle}
\end{equation}

When the active obstacle $j_t^\star$ lies within the forward camera field of view, the latest valid scene-level VLM score refreshes the semantic state associated with that interaction. If the same obstacle remains active but temporarily leaves the camera view as the robot maneuvers around it, the most recent valid score is retained while the interaction remains in the robot's forward area. When the active obstacle changes, or the previous interaction passes behind the robot, the stored semantic state is released and newly updated. 

We denote the resulting applied semantic score by $s_{\mathrm{app}}^{t}$. This procedure does not require the VLM to identify, localize, or label individual objects. The VLM continues to produce a single caution estimate, while geometric state determines which ongoing interaction that estimate informs. 

\subsection{Semantic Adaptation of CBF Conservativeness}
\label{sec:semantic_alpha}

The applied contextual caution score $s_{\mathrm{app}}^{t}$ is used to adapt the CBF parameter $\alpha$.
To emphasize differences within the operating range of the VLM signal while retaining a smooth monotone mapping, we first transform the applied caution score using a normalized logistic function. Let:

\begin{equation}
\sigma(s;k,c)=\frac{1}{1+\exp[-k(s-c)]}.
\end{equation}

We normalize the logistic response over the unit interval: 

\begin{equation}
\tilde{s}_t
=
\operatorname{clip}\!\left(
\frac{\sigma(s_t^{\mathrm{app}};k,c)-\sigma(0;k,c)}
{\sigma(1;k,c)-\sigma(0;k,c)},
0,1
\right), 
\end{equation}

where we use $k=8$ and $c=0.45$ and $\text{clip}(x,\ell,u)=\min\{u,\max\{\ell,x\}\}$ enforces bounds. Additional mapping formulations and tuning evaluations are provided in the supplementary material. Rather than mapping semantics to an unconstrained global value, \framework{} instead uses the current geometric cap as the upper bound of the semantic adaptation: 

\begin{equation}
\alpha_t
=
\operatorname{clip}\!\left(
\alpha_f+
\left(\alpha_{\mathrm{cap},t}-\alpha_f\right)
\left(1-\tilde{s}_t\right),
\alpha_{\min},
\alpha_{\mathrm{cap},t}
\right). 
\label{eq:alpha_mapping}
\end{equation}

We use the mapping parameter $\alpha_f=0.05$, with the controller enforcing its admissible lower bound $\alpha_{\min}=0.10$. This monotonic mapping means that increasing contextual caution decreases $\alpha$ and thus increases controller conservativeness. Conversely, when semantic caution is low, $\alpha$ approaches the geometric cap. Thus, by construction:

\begin{equation}
\alpha_{\min}
\leq
\alpha_t
\leq
\alpha_{\mathrm{cap},t}.
\label{eq:alpha_range}
\end{equation}

Such a formulation means that geometry determines how aggressive the controller is allowed to be, while semantics determines how much additional margin should be allocated within that range.

\subsection{Geometry- and Speed-Aware Cap}
\label{sec:dynamic_cap}

Because VLM outputs may be delayed or faulty, we instead impose a geometry- and speed-dependent upper bound on $\alpha$. The cap limits how aggressive the controller may become near obstacles, and thus acts as a geometry-only fallback whenever a valid semantic estimate is unavailable. 

Crucially, the cap is intentionally semantic-invariant. Such a design is intentional to guarantee a geometry only fallback. 

\paragraph{Distance-based cap}

For the closest obstacle, let $m_t$ denote the current clearance between the robot and the corresponding obstacle. We first normalize the clearance as: 

\begin{equation}
q_t
=
\min
\left(
1,\,
\frac{m_t}{M_{\mathrm{safe}}}
\right),
\label{eq:normalized_clearance}
\end{equation}

where $M_{\mathrm{safe}}>0$ is a clearance scale that defines when we consider the robot safely far. We then smoothly interpolate between a conservative near obstacle value $\alpha_{\mathrm{near}}$ and a more permissive far value $\alpha_{\mathrm{far}}$, with $\gamma_c>0$ shaping the transition:

\begin{equation}
\alpha_{\mathrm{dist},t}
=
\alpha_{\mathrm{near}}
+
\left(
\alpha_{\mathrm{far}}
-
\alpha_{\mathrm{near}}
\right)
q_t^{\gamma_c}.
\label{eq:distance_cap}
\end{equation}

where $M_{\mathrm{safe}}=1.5~\mathrm{m}$, $\alpha_{\mathrm{near}}=0.20$, $\alpha_{\mathrm{far}}=0.60$, and $\gamma_c=2$.

\paragraph{Speed-based scaling}

Clearance alone is not sufficient: for the same geometric margin $m$, a faster robot should permit less aggressive behavior. Let $v_t$ denote the current forward speed, and let $V_{\max}>0$ be the maximum allowed forward speed. We apply a speed-based scaling factor to $\alpha_{\text{dist}}$:

\begin{equation}
\eta(v_t)
=
\frac{1}{
1+
G_v
\left(
\frac{\max(0,v_t)}{V_{\max}}
\right)
},
\label{eq:speed_scaling}
\end{equation}

with $G_v=1$ and $V_{\max}=0.50~\mathrm{m/s}$. The resulting geometric cap is: 

\begin{equation}
\alpha_{\mathrm{cap},t}
=
\operatorname{clip}
\left(
\alpha_{\mathrm{dist},t}\eta(v_t),
\alpha_{\mathrm{near}},
\alpha_{\mathrm{far}}
\right).
\label{eq:dynamic_cap}
\end{equation}


\subsection{Asynchronous Semantic Updates}
\label{sec:async}

VLM inference is executed asynchronously in a background worker independently of the low-level control loop to avoid blocking the control loop and enable real-time robot navigation. Every $N=5$ control steps, the current RGB frame is captured and sent to the VLM, initiating a VLM semantic query while the robot continues navigating using the most recently accepted applied semantic estimate. 


Let $t_{\mathrm{vlm}}$ denote the time at which the most recent semantic result became available to the controller. We define its age at control time $t$ as: 

\begin{equation}
\tau_t
=
t-t_{\mathrm{vlm}}.
\label{eq:semantic_age}
\end{equation}

The semantic estimate is considered stale when:

\begin{equation}
\operatorname{stale}(t)
=
\mathbb{I}
\left[
\tau_t>T_{\mathrm{stale}}
\right],
\label{eq:stale}
\end{equation}

where $T_{\mathrm{stale}}=1.8~\mathrm{s}$. If no semantic estimate has yet been received, or if the VLM result exceeds the staleness threshold, semantic adaptation is disabled and the controller falls back to the geometric cap alone. Interaction persistence does not override this staleness safeguard. It preserves valid semantic context across temporary changes in viewpoint, but an estimate that becomes temporally stale is still discarded.

\section{Experiments}
\label{sec:experiments}

We evaluate \framework{} to determine whether visual semantics provide useful information for adapting CBF conservativeness beyond what can be obtained from geometry. 


\subsection{Experimental Setup and Evaluation Design}
\label{sec:exp_setup}

We evaluate \framework{} primarily in NVIDIA IsaacSim~\cite{isaacsim_repo} using a Carter differential-drive mobile robot with an egocentric RGB camera. 
Each rollout terminates upon reaching the goal, on physical contact with an obstacle, or after a maximum of $1800$ simulation steps ($30.06$ seconds). 





We compare \framework{} against the following ablations and baselines:

\begin{itemize}

\item \emph{Fixed $\alpha$:} 
We evaluate a sweep of constant $\alpha$ values from permissive to conservative. This establishes the tradeoff obtainable by globally changing controller conservativeness without any semantic adaptation.

\item \emph{Geometry-only adaptation:} 
We remove the semantic VLM pathway and directly use the geometric cap, $\alpha_t=\alpha_{\mathrm{cap},t}$. This retains the same geometric adaptation used by \framework{} while removing visual semantics.

\item \emph{Online Adaptive CBF (OA-CBF):~\cite{kim2026learning}}
We compare against an adaptive-CBF method that changes its $\alpha$ parameter online, via a learned policy, without access to the VLM semantic signal. 

\item \emph{Parameter Adaptive CBF (PACBF):~\cite{kim2025adaptcontrolbarrierfunctions}}
We additionally compare against a parameter-adaptive CBF implementation. PACBF uses online estimators to adapt its barrier function parameters from geometric and dynamic state, but receives no visual semantic input.

\end{itemize}




\subsection{Contextual Caution Signal Analysis}
\label{sec:vlm_analysis}

\begin{figure*}[t]
    \centering
    \includegraphics[scale=0.48]{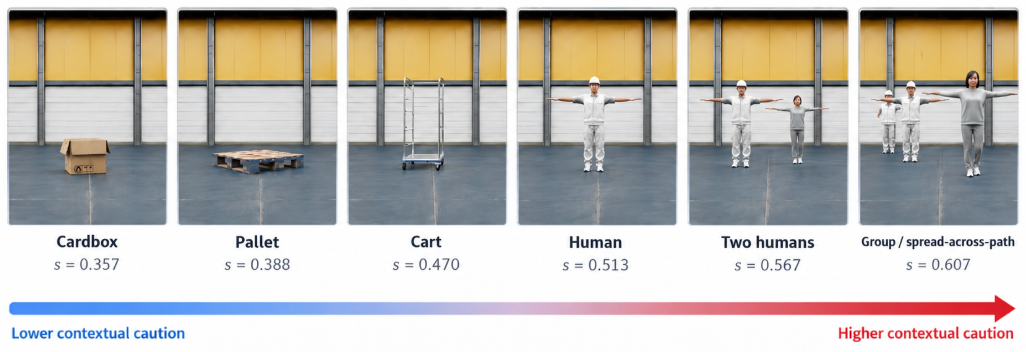}
    \caption{VLM-derived contextual caution scores, least to highest.}
    \label{fig:semantic_scale}
    \vspace{-18pt}
\end{figure*}

We first examine the VLM signal independently of closed-loop navigation. To isolate visual semantics, we render a variety of different entities at the same target pose while keeping the modeled obstacle radius and surrounding geometry fixed. 

The contextual-caution score varies substantially with visual identity: 0.357 for a cardboard box, 0.388 for a warehouse pallet, 0.470 for a cart, 0.479 for a barrel, 0.513 for a human, and 0.553 for a hospital bed. The resulting ordering indicates that the signal is not a binary human detector. Rather, it captures broader contextual cues about whether the robot should yield to or avoid crowding the visible entity. When further varying social context, scores increase from 0.513 for a single human to 0.567 for two humans, and reach 0.607 when 3 humans are distributed across the robot's forward path. A compact three person clutter produces 0.544, indicating that caution depends on scene configuration rather than simply the number of visible people. The robot scene images can be observed in Fig.~\ref{fig:semantic_scale}. The ordering reflects the generality of the prompt. The VLM is asked how strongly the robot should yield to whatever is visible. Thus, movable equipment, like a hospital bed, also rightfully received elevated caution. 

Additionally, to examine the potential headroom of the semantic pathway with stronger multimodal models, we query GPT-5.6 offline on egocentric observations also used with our Qwen2.5-VL-7B backbone. On five clean matched human--object pairs, GPT-5.6 produces a substantially larger mean human--object score gap than Qwen (0.80 vs 0.14), while preserving the same ordering on all five pairs. This suggests that stronger multimodal models may provide more discriminative semantic signals without requiring changes to \framework{}'s downstream control architecture. Current frontier models do not yet meet the latency requirements of our online system, motivating our use of a lightweight local VLM in the present implementation. As multimodal inference becomes more efficient, \framework{}'s modular semantic interface allows stronger backbones to be incorporated without changing the underlying adaptation mechanism. 

\subsection{Matched Geometry Semantic Selectivity}
\label{sec:matched_eval}

We next test whether semantic information produces a measurable change in navigation behavior when geometry is held fixed. We construct paired scenes with identical robot start, goal, and obstacle configuration, changing only the semantic identity of the target. 

Under the geometry-only controller, changing the rendered identity does not affect the trajectory. Both dynamic CBF baselines, OA-CBF and PACBF are likewise invariant to the visual swap. On the exact same evaluation geometry, all three non-visual methods produce a human--object margin differential of 0 m. In contrast, \framework{} produces a semantic differential of 0.405 m more margin around the human than around the benign object. 

Importantly, this additional human reserve does not require applying the same conservativeness globally. The benign object path length under \framework{} is 9.413 m, only 1.4\% longer than the 9.282 m geometry-only path. By comparison a globally fixed conservative $\alpha$ produces a similar margin difference to \framework{}, but incurs a 5.0\% benign object case path overhead. Thus, achieving comparable reserve through global conservativeness costs roughly 3.5 times that of \framework{}.


\begin{table}[t]
\centering
\caption{Matched-geometry semantic selectivity. Modeled target margin is measured under identical physical configurations while only the rendered target identity is changed. \framework{} successfully differentiates between a human and a benign object and leaves greater clearance for the human, while the others are purely geometric and thus output the exact same trajectories.}
\label{tab:matched_semantic}
\resizebox{\columnwidth}{!}{
\begin{tabular}{rccc}
\toprule
Method & \makecell{Object \\ margin (m)} & \makecell{Human \\ margin (m)} & \makecell{$\Delta$ \\ margin (m)} \\
\midrule
OA-CBF~\cite{kim2026learning}        & 0.688 & 0.688 & 0.000 \\
PACBF~\cite{kim2025adaptcontrolbarrierfunctions}         & 1.052 & 1.052 & 0.000 \\
Geometry-only cap & 0.963 & 0.963 & 0.000 \\
\textbf{\framework{}} & \textbf{1.141} & \textbf{1.546} & \textbf{+0.405} \\
\bottomrule
\end{tabular}}
\vspace{-18pt}

\end{table}


\subsection{Dynamic Human Robustness in Increasingly Complex Scenes}


Our previous experiment shows that VLM semantics create additional behavioral reserve around a static human. We next test whether that reserve has downstream value when the environment subsequently changes and the human moves. We first consider continuous human motion, varying the speed of a human continuously moving across the robot's path. The geometry-only controller leads to physical collisions over a contiguous range of tested human speeds from 0.7 to 0.9 m/s. In contrast, \framework{} shifts this collision boundary, remaining completely collision-free through the 0.7-0.85 m/s interval. 



\begin{figure}[t]
    \centering
    \includegraphics[width=\columnwidth]{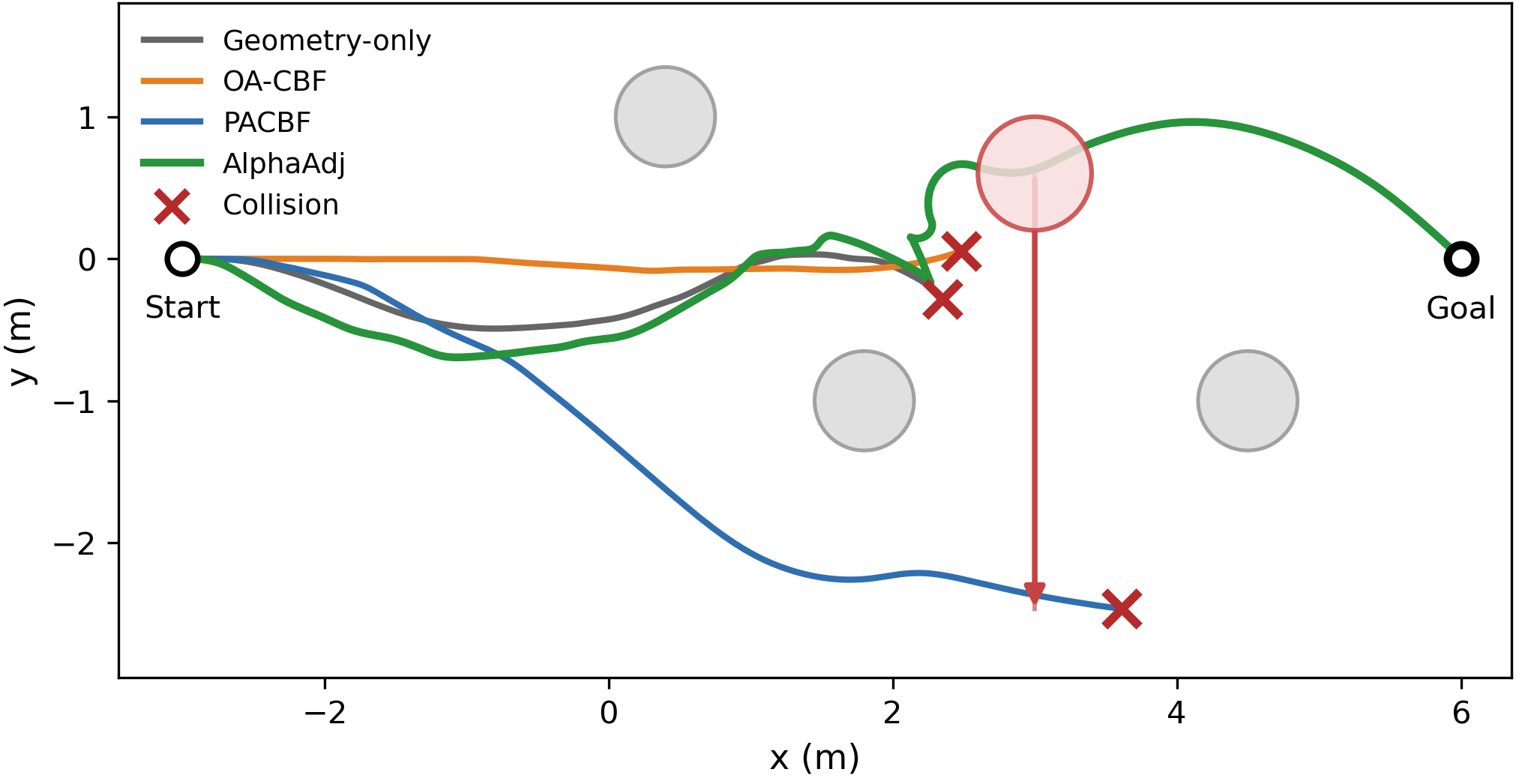}
    \caption{
    Dynamic human robustness under delayed, unexpected motion in the cluttered stress test. \framework{} remains collision-free while geometry-only, OA-CBF, and PACBF all collide with the human.
    }
    \label{fig:dynamic_human_robustness}
    \vspace{-18pt}
\end{figure}

We next consider delayed, unexpected human motion. Each trial begins with a stationary human. The VLM receives only the current RGB image and is not informed that the human will later move. After the robot has observed the static scene and begun navigating, the human begins moving according to a predetermined perturbation. Across the resulting sweep, we identify multiple conditions in which the additional reserve provided by \framework{} prevents contact when all three geometric baselines fail. For example, when the human begins moving at 0.4 m/s once the robot reaches a position of 1.7m away from the start, \framework{} succeeds in all trials, whereas the geometric cap controller collides and OA-CBF collides in all trials. 

Finally, we repeat the delayed-motion test in a more challenging multi-obstacle warehouse environment. Three additional CBF obstacles are arranged sequentially around the semantic target, requiring the robot to negotiate several local interactions rather than avoid a single isolated obstacle. In the static cluttered scene, all evaluated methods reach the goal without physical contact, while \framework{} retains similar path geometry to the geometry-only controller: its object and human paths are only 2.1\% and 2.5\% longer, respectively. That small increase in path length pays off when delayed human motion is introduced into this same cluttered scene. \framework{} reaches the goal without physical contact in all trials. In contrast, geometry-only cap, OA-CBF, and PACBF each incur true physical contact in all trials under the identical scene and human motion. Fig.~\ref{fig:dynamic_human_robustness} illustrates the resulting trajectories. Despite its semantic caution, \framework{} initially makes relatively direct progress toward the goal, only taking a slightly wider berth. Once the delayed human motion creates an imminent interaction, however, its trajectory bends around the moving human and continues to the goal without collision. PACBF takes a substantially wider berth much earlier in the trajectory, but still collides under the delayed motion. Geometry-only and OA-CBF also make physical contact. This behavior highlights the selective nature of \framework{}'s semantic reserve. Additional conservativeness is expressed when the interaction demands it rather than uniformly along the entire trajectory. Together with the clean dynamic experiments, these results show that the benefit of semantic reserve persists beyond isolated interactions and into more cluttered environments.

Altogether, these experiments show that the semantic reserve is not merely a change in static trajectory geometry. It shifts the range of human motions the robot can accommodate without physical contact, both in isolated interactions and in more complex multi-obstacle scenes. \framework{} does not explicitly predict future human motion. Instead, VLM-derived semantics induce behavior that allows the robot to better respond to sudden movements and scene changes.


This result also highlights the safety-efficiency tradeoff induced by \framework{}'s semantic reserve. In the static version of the cluttered scene, \framework{} sacrifices a slight amount of path efficiency, but under unpredictable human movement, it is the only evaluated method to remain collision-free.

\subsection{Real-Time Asynchronous Performance}
\label{sec:experiments_async}

\framework{} decouples VLM inference from the low-level control loop, allowing control to proceed continuously while semantic estimates are computed in the background. Across 280 VLM calls in a representative run, mean end-to-end VLM round-trip latency is 0.158 seconds.
This latency is already substantially lower than the VLM-side latency reported by prior VLM-navigation systems, as seen in Table~\ref{tab:runtime_comparison}. For example, VLM-Nav~\cite{goetting2024endtoendnavigationvisionlanguage} reports an average VLM-side latency of 9.07 seconds, while VLM-Social-Nav~\cite{song2024vlmsocialnavsociallyawarerobot} reports 1.75 seconds. Our direct token-probability readout avoids waiting for full autoregressive response generation, enabling semantic updates on a substantially shorter timescale. 

More importantly, \framework{} goes a step further by removing VLM inference from the critical control path altogether. The low-level controller runs continuously with one control step every $0.0167$ s, using the most recent valid applied semantic state while new VLM estimates are computed asynchronously. 
In synchronous VLM-based approaches, the VLM response time directly constrains how quickly each decision can be produced. Beyond the  VLM-side latencies reported above, synchronous Vision--LLM in~\cite{jmse13081553} reports an overall decision--control loop time of approximately $6.2$ seconds. In contrast, \framework{} combines fast semantic inference with asynchronous execution, allowing the controller to maintain its 60 Hz update rate independently of VLM response time. For context, OA-CBF and PACBF both use control periods of $0.05$ seconds (20 Hz). Thus \framework{} preserves a high-rate control loop while incorporating semantic guidance on an independently scheduled timescale. 

\begin{table}[t]
\centering
\caption{Latency and execution characteristics of VLM-based navigation approaches. 
\framework{} combines low-latency semantic inference with asynchronous execution, allowing control to continue independently of VLM response time.}
\label{tab:runtime_comparison}
\resizebox{\columnwidth}{!}{
\begin{tabular}{rcc}
\toprule
Method & \makecell{Reported VLM (s) \\ (loop latency)} & Execution \\
\midrule
VLM-Nav~\cite{goetting2024endtoendnavigationvisionlanguage}
    & 9.070 & Synchronous \\
VLM-Social-Nav~\cite{song2024vlmsocialnavsociallyawarerobot}
    & 1.750 & Synchronous \\
Vision--LLM~\cite{jmse13081553}
    & \makecell{6.200 \\(decision--control)} & Synchronous \\
\midrule
\framework{} VLM
    & 0.158 & Asynchronous \\
\framework{} control
    & \textbf{0.0167} & \textbf{Continuous} \\
\bottomrule
\end{tabular}}
\vspace{-18pt}
\end{table}

\subsection{Physical Robot}

\begin{figure}[t]
    \centering
    \includegraphics[width=\columnwidth]{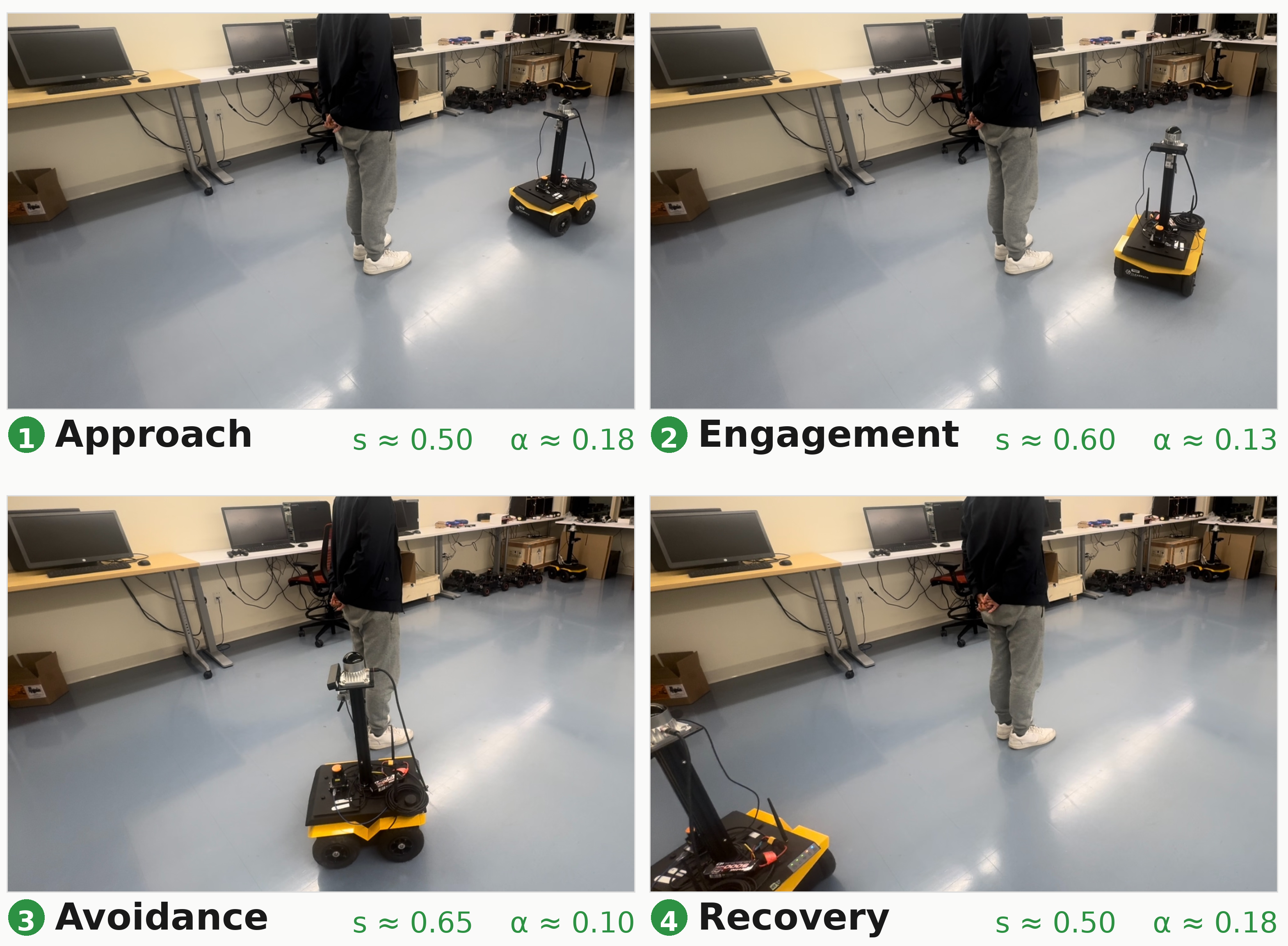}
    \caption{
    Real-robot implementation of \framework{} on a Clearpath Jackal during a controlled human encounter.
    The robot approaches, adapts its semantic caution and CBF parameter $\alpha$, executes an avoidance maneuver, and recovers after passing the person.
    }
    \label{fig:jackal}
    \vspace{-19pt}
\end{figure}

We deploy \framework{} on a Clearpath Jackal equipped with an egocentric RGB camera and geometric obstacle sensing. Fig.~\ref{fig:jackal} shows representative stages of the human encounter. The physical system preserves the simulation pipeline: the VLM produces an online semantic caution score, which adapts the CBF parameter $\alpha$, while the sampled-CBF controller replans at every control step. 

As the robot approaches the person, semantic caution increases and $\alpha$ decreases, resulting in a more conservative control response and a clear avoidance maneuver before the robot proceeds toward its goal. This demonstrates that the semantic adaptation and closed-loop behavior of \framework{} transfer to a physical mobile robot rather than being specific to the Isaac Sim environment.



\section{Conclusion, Limitations, and Future Work}
\label{Conclusion}
In this work, we introduced \framework{}, a framework for using VLM-derived visual semantics to adapt the conservativeness of a CBF safety filter in real time robot navigation. Our experiments show that this semantic information produces strong and robust behavior that geometry alone cannot. Taken together, the results support a view of robot control where geometry determines the robot's physical constraints, while VLM semantics give additional insights to when additional behavioral reserve is required. 

A limitation of \framework{} is that its effectiveness depends on the quality and update rate of the underlying VLM. With lightweight VLMs, semantic separation can slowly weaken as scenes get more complex. Future work can further investigate stronger VLMs, richer semantic representations beyond a scalar, and larger scale real world evaluation.







\bibliographystyle{IEEEtran} 
\bibliography{refs}
\flushbottom

\end{document}